\documentclass{article}

\usepackage[preprint]{neurips_2025}

\usepackage[utf8]{inputenc} 
\usepackage[T1]{fontenc}    
\usepackage{hyperref}       
\usepackage{url}            
\usepackage{booktabs}       
\usepackage{amsfonts}       
\usepackage{nicefrac}       
\usepackage{microtype}      
\usepackage{xcolor}         
\usepackage{algorithm}
\usepackage{algpseudocode}
\usepackage{enumitem}
\usepackage{mdframed}

\usepackage{amsmath}
\usepackage{amssymb}
\usepackage{mathtools}
\usepackage{amsthm}
\usepackage{bbm}
\usepackage{mathtools}

\usepackage[capitalize,noabbrev]{cleveref}

\theoremstyle{plain}
\newtheorem{theorem}{Theorem}[section]
\newtheorem{proposition}[theorem]{Proposition}

\theoremstyle{definition}
\newtheorem{definition}[theorem]{Definition}

\theoremstyle{remark}
\newtheorem{remark}[theorem]{Remark}

\author{
  Yinglun Xu\thanks{Equal contribution},~
  Hangoo Kang\footnotemark[1],~
  Tarun Suresh,~
  Yuxuan Wan,~
  Gagandeep Singh \\
  University of Illinois Urbana-Champaign\\
  \texttt{\{yinglun6, hangook2, tsuresh3, yuxuanw8, ggnds\}@illinois.edu}
}

\title{Learning a Pessimistic Reward Model in RLHF}

\begin{document}

\maketitle

\begin{abstract}
    This work proposes `PET', a novel pessimistic reward fine-tuning method, to learn a pessimistic reward model robust against reward hacking in offline reinforcement learning from human feedback (RLHF). Traditional reward modeling techniques in RLHF train an imperfect reward model, on which a KL regularization plays a pivotal role in mitigating reward hacking when optimizing a policy. Such an intuition-based method still suffers from reward hacking, and the policies with large KL divergence from the dataset distribution are excluded during learning. In contrast, we show that when optimizing a policy on a pessimistic reward model fine-tuned through PET, reward hacking can be prevented without relying on any regularization. We test our methods on the standard TL;DR summarization dataset. We find that one can learn a high-quality policy on our pessimistic reward without using any regularization. Such a policy has a high KL divergence from the dataset distribution while having high performance in practice. In summary, our work shows the feasibility of learning a pessimistic reward model against reward hacking. The agent can greedily search for the policy with a high pessimistic reward without suffering from reward hacking.
\end{abstract}
\section{Introduction}\label{sec:intro}
Reinforcement learning from human feedback (RLHF) 
\citep{christiano2017deep} has become crucial in aligning large language models (LLMs), making LLMs more helpful, truthful, and harmless \citep{stiennon2020learning,bai2022training,ouyang2022training,rafailov2023direct}. In a typical RLHF training framework \citep{ouyang2022training}, an agent first learns a reward model as a proxy for human preference that best interprets the preference training data. Then, the agent optimizes a policy on the proxy reward model to achieve a high reward. In the real world, such a standard training framework faces the critical problem of `reward hacking,' also known as `reward over-optimization' \citep{eisenstein2023helping,tien2022causal,gao2023scaling}. Due to the dataset's size limitation, the proxy reward model is not always accurate. Given a prompt (task input), there could exist a response (task output) that is not favored by human preference, but the learned proxy reward model overestimates the response and gives it a high reward. A greedy agent that searches for the policy with the highest proxy reward can learn to output such low-quality and overestimated responses. This is a typical instance of reward hacking. Due to the over-estimations in the proxy reward model, reward hacking can happen during policy optimization and decrease the learning efficiency of an algorithm \citep{gao2023scaling}, making it one of the most urgent challenges in RLHF. 

\textbf{Challenges in preventing reward hacking:}
To avoid reward hacking, the idea of `pessimism' is necessary for the learning agent \citep{levine2020offline}. At a high level, the agent must be pessimistic when evaluating the policy's performance based on the dataset. For example, in the framework described above, to avoid over-estimation, the agent should expect the case where the true performance of a policy is less than its performance evaluated based on the proxy reward. The key challenge here is controlling the degree of pessimism, that is, how low the true performance of a policy can be based on the preference dataset. While sufficient pessimism can avoid over-estimation and mitigate reward hacking, over-pessimism results in an under-estimation of high-quality policies, causing the agent to exclude these good policies during training. 

KL regularization plays a pivotal role in current RLHF methods to mitigate the risk of reward hacking \cite{ouyang2022training}. Intuitively, the KL divergence between the response distributions of a policy and the dataset can indicate the degree of uncertainty in evaluating the policy. A policy with higher uncertainty in evaluation requires the agent to be more pessimistic about it. Current algorithms explicitly (such as PPO \cite{schulman2017proximal}) or implicitly (such as rejection sampling \cite{touvron2023llama}) rely on the KL divergence during policy optimization to mitigate the over-estimations in the proxy reward. However, such intuition-based methods are not always efficient in controlling pessimism. Reward hacking is still observed when using such methods \cite{gao2023scaling,rafailov2024scaling}, indicating that they are not pessimistic enough for some policies. In the meantime, the methods may be overly pessimistic for some policies with a large KL divergence from the dataset distribution. This is verified by our empirical observations in Section \ref{sec:exp}. Recent studies use the adversarial training technique \cite{zhang2024self,xie2024exploratory,liu2024provably,cen2024value, ji2024self} to achieve provable robustness against reward hacking. However, these methods are based on the direct policy optimization (DPO) technique \cite{rafailov2023direct} that requires KL regularization, which again unnecessarily induces over-pessimism. While KL regularization has almost always been considered necessary in practice, it is unlikely to be the most efficient approach to control pessimism. Therefore, a research question arises:

\textit{\textbf{Can efficient pessimism be achieved by learning a pessimistic reward model whose prediction is trustworthy? Can an agent greedily optimize a policy on the pessimistic reward without using regularization to prevent reward hacking?}}

\begin{figure*}[!t]
    \centering
    \includegraphics[width=0.75\linewidth]{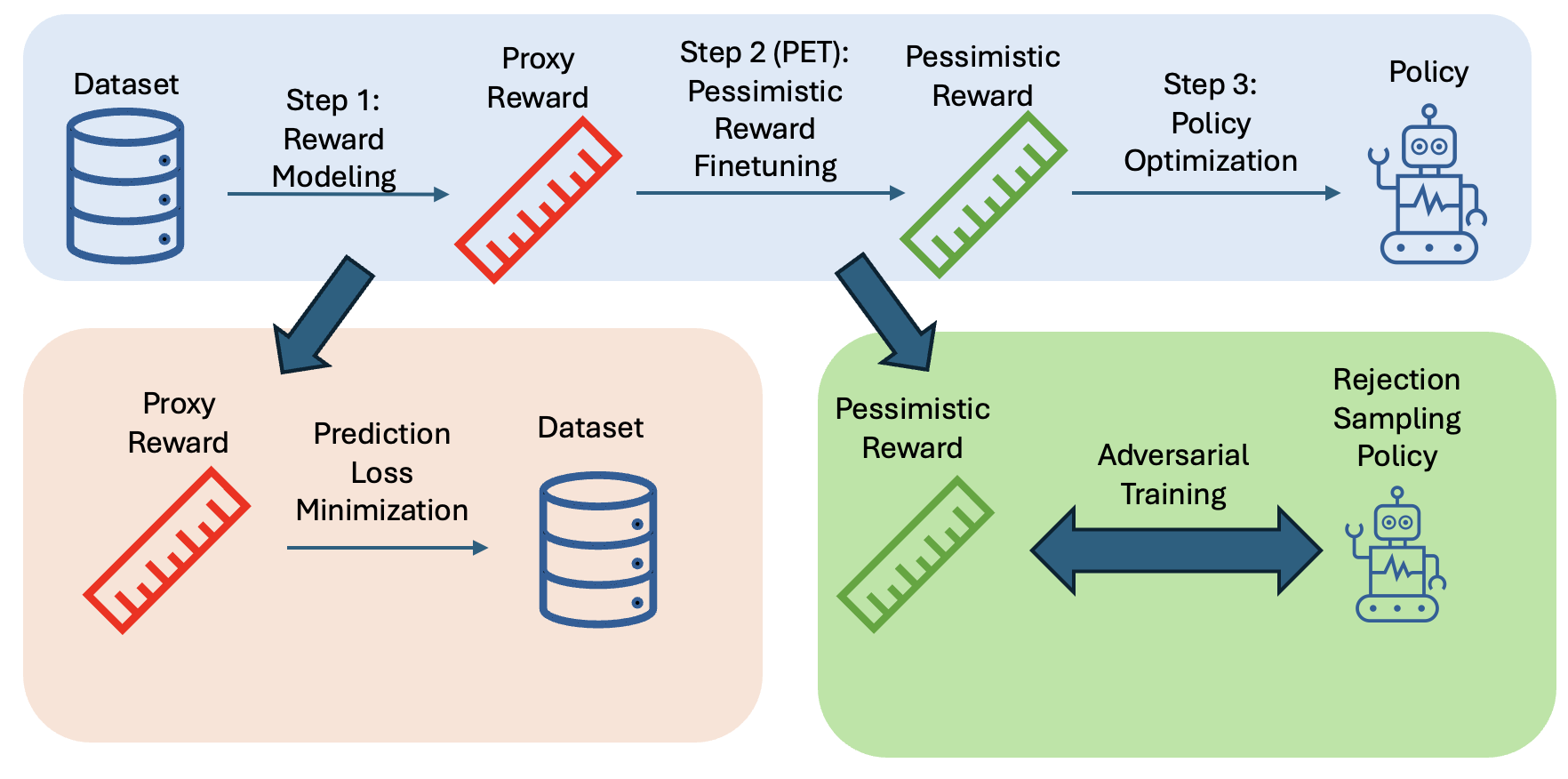} \\
    \caption{A three-step reward-based learning framework. The first step is the traditional reward modeling that trains a reward model with minimal loss on predicting the dataset preference. The second step fine-tunes the learned reward model to make it pessimistic. Particularly, the reward model is adversarially trained against a policy model induced by the rejection sampling process. The reward model should still induce minimal prediction loss on the dataset. In the last step, the framework optimizes a policy on the pessimistic reward and outputs the learned policy.}
    \label{fig:intro}
    \vspace{-4mm}
\end{figure*}

\textbf{Rejection sampling with adversarial training:}
In this work, we provide positive answers to the above questions. Our method is based on the adversarial training technique \citep{cheng2022adversarially,bhardwaj2023adversarial,zhan2023provable, gupta2025mitigating}. In the framework, a policy model is trained to achieve a high reward on a reward model, while the reward model is trained to give a low reward to the policy model. Here, the main challenge comes from the instability of adversarial training. Intuitively, the policy model is optimized on different reward models during training, which complicates training convergence. Current works, such as \cite{liu2024provably}, mitigate the challenge by using the DPO technique to simplify the training process. However, as discussed earlier, such a method requires KL regularization, which induces over-pessimism. To mitigate the challenge without using regularization, we consider using the rejection sampling technique \citep{touvron2023llama,beirami2024theoretical} to simplify the adversarial training process. Given an initial policy model and a reward model, the rejection sampling process samples several responses from the policy model and outputs the response with the highest reward as described in Alg \ref{alg:bon} in the appendix. Rejection sampling performs policy optimization at inference time, making it easy to implement and compute. Note that the traditional rejection sampling process suffers from reward hacking on the proxy reward \cite{touvron2023llama}. Fortunately, the combination of rejection sampling with adversarial training simultaneously solves the problems of reward hacking in rejection sampling and the difficulty of policy optimization in adversarial training. Based on the idea of adversarial training against rejection sampling, we develop a novel reward fine-tuning method and a corresponding three-step RLHF algorithm. More specifically, our contributions are as follows:

\begin{enumerate}[leftmargin=*]
    \item We develop a novel reward fine-tuning method called `PET'. A reward model fine-tuned by PET is pessimistic and robust against reward hacking. Under standard assumptions \cite{liu2024provably}, we theoretically prove that the rejection sampling process on the pessimistic reward fine-tuned by PET has a performance comparable to the rejection sampling process on any reward, as long as the corresponding policy induced by the process is covered by the dataset.
    \item Based on `PET', we develop a three-step RLHF framework. In Fig. \ref{fig:intro} we intuitively show how the framework works. The first step is the standard reward modeling that learns a proxy reward model with minimal prediction loss on the preference dataset. The second step is PET, which fine-tunes the proxy reward to make it pessimistic. The last step is policy optimization on the learned pessimistic reward model. In principle, in the last step, the agent can greedily optimize a policy to achieve the highest reward on the pessimistic reward model without using any regularization.
    \item We test our method on the TL;DR summarization dataset \cite{volske2017tl} and the IMDB dataset \cite{maas-etal-2011-learning}. On the summarization dataset, we observe that the rejection sampling process on a pessimistic reward fine-tuned by PET significantly outperforms the traditional sampling process on the proxy reward. The win rate of the policy response against human response is increased from $32.0\%$ to $39.2\%$. We further use the PPO algorithm for policy optimization, which we call `PPO-PET'. We observe that the policy learned by PPO-PET has a high performance and a high KL divergence from the dataset distribution. This contradicts the traditional impression that policies with high KL divergence are vulnerable to reward hacking. On the summarization dataset, our policy in general has a comparable or higher performance compared to current state-of-the-art RLHF algorithms, including DPO \cite{rafailov2023direct}, RPO \cite{liu2024provably}, and $\chi$PO \cite{huang2024correcting}. On the IMDB dataset,  the ground truth reward is available, and the policy learned by PPO-PET achieves a reward significantly higher than the baselines.
\end{enumerate}

\section{Preliminaries}\label{sec:prelim}
\subsection{Offline RLHF}
In this work, we consider a standard RLHF problem in the offline learning setting. For a given task, let $\mathcal{X}$ be the input space (e.g., prompt), and $\mathcal{A}$ be the output space (e.g., response). There exists a preference model $\mathcal{P}(\cdot|\cdot,\cdot,\cdot):\mathcal{X}\times\mathcal{A}\times\mathcal{A}\rightarrow\Delta(\{\succ,\prec\})$ that takes a prompt $x$ and a pair of responses $a_1,a_2$ as input and stochastically outputs a preference signal $\sigma \sim \mathcal{P}(\cdot|x,a_1,a_2)$, representing its preference on the two responses. For example, $\sigma$ being $\succ$ means the preference is $a_1\succ a_2$ given the prompt $x$. In addition, we assume the preference model is a Bradley-Terry model \cite{bradley1952rank} associated with a reward model $r:\mathcal{X} \times \mathcal{A} \rightarrow \mathbb{R}$ as $\mathcal{P}_{r}(\succ|x,a_1,a_2)=\text{sigmoid}(r(x,a_1)-r(x,a_2)),$ where $\text{sigmoid}(y):=\frac{1}{1-\exp(-y)}$ is the sigmoid function.

Here, the reward model $r$ represents the quality of a response. Given a prompt, a response with a higher reward is more likely to be preferred. We say $r$ is the true reward behind the preference model $\mathcal{P}_{r}$. A policy $\pi(\cdot|\cdot):\mathcal{X}\rightarrow \Delta(\mathcal{A})$ takes a prompt $x$ as an input and stochastically outputs a response $a \sim \pi(\cdot|x)$. Let $\mu$ be a probability distribution over the prompt space, the quality of the policy on the reward $r$ over the distribution $\mu$ is represented by $V^r_\mu(\pi)=\mathbb{E}_{x\sim \mu,a\sim \pi(\cdot|x)}[r(x, a)].$

In the offline RLHF setting, there exists an offline preference dataset consisting of $N$ tuples $\mathcal{D}=\{(x^i,a_1^i,a_2^i,\sigma^i)\}_{i=1}^N.$ The prompt and responses in each tuple is sampled from an i.i.d. distribution $(x^i,a_1^i,a_2^i) \sim \mu_\mathcal{D}$, and the preference signal is sampled from the preference model $\sigma^i \sim \mathcal{P}(\cdot|x^i,a_1^i,a_2^i).$ We call $\mu_D$ the dataset distribution and denote $\mu$ as the distribution of the prompts in $\mu_{\mathcal{D}}$. The agent has access to the offline dataset $\mathcal{D}$, the prompt space $\mathcal{X}$, the response space $\mathcal{A}$, and the prompt distribution $\mu$. Let $r^*$ be the true reward model behind the preference model that generates the preferences in the dataset. The general goal of the agent is to align a policy with the reward model reliably, that is, to find a policy of high performance $V^{\mu}_{r^*}(\pi)$ with a high probability.

\subsection{Reward Modeling}
Reward modeling is a process to learn a reward model that can well explain the preference signals in the dataset. Given a reward model $\hat{r}$, its prediction loss on the preference dataset $\mathcal{D}$ is defined as $\mathcal{L}_{\mathcal{D}}(\hat{r})=\sum_{i=1}^N - \log \mathcal{P}_{\hat{r}} (\sigma^i|x^i,a_1^i,a_2^i).$ The prediction loss of a reward model represents its quality in interpreting the preference signals in the dataset. The process of traditional reward modeling is to find the reward model from a model class $\mathcal{R}$ (e.g., an LLM) that minimizes the prediction loss over the offline dataset: $\hat{r} \in \arg\min_{r \in \mathcal{R}}\mathcal{L}_{\mathcal{D}}(r).$ For convenience, we say a reward model is a `proxy reward model' if it has low prediction loss on the dataset.

\subsection{Policy Optimization}
After learning a reward model, the next step is to optimize a policy on the reward model to achieve a high reward. In the appendix, we provide a detailed description of two typical policy optimization techniques most related to our study: KL regularized proximal policy optimization (KL-PPO) and rejection sampling (RS). Here, we briefly go over the process of rejection sampling. Rejection sampling, also known as best-of-N sampling, is an inference-time policy optimization method. Given a base policy model $\pi_0$, a reward model $\hat{r}$, and a positive integer $n$, the process of rejection sampling is defined in Alg \ref{alg:bon} in the appendix. In practice, the reward model $\hat{r}$ is trained by minimizing the prediction loss, and the base policy $\pi_0$ is usually set as a proxy policy for the dataset. The proxy policy is trained to reproduce the response in the dataset. This is often achieved by a standard supervised fine-tuning process on the prompts and responses from the dataset $\mathcal{D}$ \citep{touvron2023llama}. The rejection sampling process is effectively a policy as it takes a prompt as input and stochastically outputs a response. We denote $\pi_{\text{RS}}(\pi_0,\hat{r},n)$ as the policy to represent the rejection sampling process, which we call the `RS policy'. Particularly, we highlight an intuitive yet important property of a rejection sampling policy in Proposition \ref{pro:rs}. The proof is in the appendix. Later, we show how this property helps us implement our method.

\begin{proposition}\label{pro:rs}
    For any prompt distribution $\mu$, base policy $\pi_0$, number of sampling $n$, and reward model $r_0$, the rejection sampling policies satisfy: $$V^\mu_{r_0}(\pi_{\text{RS}}(\pi_0,r_0,n)) = \max_{r}V^\mu_{r_0}(\pi_{\text{RS}}(\pi_0,r,n)).$$
\end{proposition}

\section{Robust RLHF with Pessimistic Reward Finetuning (PET)}\label{sec:method}



\textbf{Robust rejection sampling as a minimax problem:}
We achieve robustness against reward hacking by training a pessimistic reward model. Specifically, we aim to find a reward model with a low prediction loss on the dataset that gives a minimal relative score to its corresponding RS policy. The relative score is compared to the score of a fixed reference policy. In practice, the reference policy is usually the proxy policy of the dataset. We focus on the relative score because the preference model only depends on the relative score between two responses. We consider RS policies because they are easy to acquire and represent near-optimal policies on a reward model. In addition, later we will show how the property in Proposition \ref{pro:rs} reduces the difficulty of implementing our algorithm.

Formally, to find such a pessimistic reward, we consider solving a minimax problem where a critic and an actor are trained against each other. The actor controls a policy model, and the critic controls a reward model. The minimax problem is as follows.

\begin{equation}\label{eq:mini}
    \min_{r \in \mathcal{R}}\max_{\pi \in \Pi^{n,\pi_0}_{\text{RS}}} \bigl(V^\mu_r(\pi)-V^\mu_r(\pi_{\text{ref}})\bigr)+\beta \cdot \mathcal{L}_{\mathcal{D}}(r).
\end{equation}

Here, $\mathcal{R}$ is a reward model class, $\Pi_{\text{RS}}^{n,\pi_0}=\{\pi_{\text{RS}}(r,n,\pi_0): r\in \mathcal{R}\}$ is the class of the RS policies with base policy $\pi_0$, $n$ samples, and any reward model $r\in \mathcal{R}$. The minimax goal $f(\pi,r):=\bigl(V^\mu_r(\pi)-V^\mu_r(\pi_{\text{ref}})\bigr)+\beta \cdot \mathcal{L}_{\mathcal{D}}(r)$ consists of two terms. The first term $\bigl(V^\mu_r(\pi)-V^\mu_r(\pi_{\text{ref}})\bigr)$ is a relative score between a policy $\pi$ and a reference policy $\pi_{\text{ref}}$ evaluated by the reward model $r$ over a prompt distribution $\mu$. The actor aims to find a policy that achieves a high score on the reward model, while the reward model aims to give a low relative score to the policy. The second term $\beta \cdot \mathcal{L}_{\mathcal{D}}(r)$ is the prediction loss on the preference dataset $\mathcal{D}$ multiplied by a positive weight $\beta > 0$, which is only determined by the reward model. This term constrains the reward model in that it must be a high-quality reward model that well explains the dataset. Such a constraint prevents the critic from being overly pessimistic. Intuitively, we want the critic to find a tight lower bound on the true performance of a policy.

\begin{remark}
    \citet{liu2024provably} proposes to solve a similar minimax problem as ours. Here, we highlight two critical differences between the formulations. The first difference is that the KL regularization is included in \citep{liu2024provably} but not in ours. Our insight is that the idea of pessimism is already included in formulating a minimax game between the policy and the reward models, so it is not necessary to add KL regularization for more pessimism. In the appendix, we show that in the theoretical analysis, the KL regularization results in over-pessimism and decreases the learning efficiency, which is undesired. Our formulation has no regularization and achieves a tighter theoretical guarantee on learning efficiency. Second difference is that we focus on the policies based on rejection sampling. Using the rejection sampling process allows us to avoid KL regularization in problem formulation. In addition, the rejection sampling process is a convenient policy optimization technique and makes the implementation easy for adversarial training, as we will show next.
\end{remark}

\textbf{PET and three-step RLHF:}
Based on the result from Proposition \ref{pro:rs}, we have $\pi_{\text{RS}}(\pi_0,n,r) \in \max_{\pi \in \Pi^{n,\pi_0}_{\text{RS}}} V^\mu_r(\pi).$ Therefore, the minimax problem in Eq \ref{eq:mini} can be simplified to a minimization problem as shown in Eq. \ref{eq:pet}.

\begin{equation}\label{eq:pet}
    \min_{r \in \mathcal{R}}\bigl(V^\mu_r(\pi_{\text{RS}}(\pi_0,n,r))-V^\mu_r(\pi_{\text{ref}})\bigr)+\beta \cdot \mathcal{L}_{\mathcal{D}}(r).
\end{equation}

To solve the minimization above, we propose a novel algorithm called `pessimistic reward fine-tuning' (PET). In the Appendix, we explain in detail that PET is essentially the process of stochastic gradient descent on the minimization goal in Eq \ref{eq:pet}. We show PET in Algorithm \ref{alg:advbon}. In PET, the policy model and the reward model are iteratively updated. In each iteration, the policy model is updated through the rejection sampling process on the current reward model. The reward loss is computed as the minimization goal in Eq. \ref{eq:pet} with the updated policy model on a batch of prompts sampled from the dataset. Then, the reward model performs a gradient descent step on the reward loss.

\begin{algorithm}[tb]
    \caption{Pessimistic Reward Fine-Tuning (PET) with Adversarially Trained Rejection Sampling}
    \begin{algorithmic}[1]
    
    \State {\bfseries Input:} Initial reward model $\hat{r}$, Dataset $\mathcal{D}$, base policy $\pi_0$, reference policy $\pi_{\text{ref}}$, number of samples $n$, pessimistic coefficient $\beta$, learning rate $\alpha$
    
    \State {\bfseries Initialize:} $r^1 \leftarrow \hat{r}$ 
    
    \For{$t=1,\ldots,T$}
    
    \State Update $\pi^t \leftarrow \pi_{\text{RS}}(\pi_0,n,r^t)$
    
    \State Sample mini batch $\mathcal{D}_{t}=\{x_i,a^1_i,a^2_i,\sigma_i\}_{i=1}^M, (x_i,a^1_i,a^2_i,\sigma_i)\overset{\text{i.i.d}}{\sim} \mathcal{D}$

    \State Sample responses $a_i\sim\pi^t(\cdot|x_i),a_{\text{ref},i}\sim \pi_{\text{ref}}(\cdot|x_i),\forall i\in[M]$
    
    \State Compute pessimistic loss $l^t=\sum_{i\in[M]} [r^t(x_i,\pi^t(x^i,a_i))-r^t(x_i,\pi^t(x^i,a_{\text{ref},i}))+\beta \cdot \mathcal{L}_{\mathcal{D}_t}(r^t)]$

    \State Update $r^{t+1}\leftarrow r^t - \alpha \cdot \nabla l^t$

    \EndFor
    
    \State {\bfseries Return:} {Reward model $r^{T+1}$}
    \end{algorithmic}
    \label{alg:advbon}
\end{algorithm}

Alg \ref{alg:advbon} proposes a pessimistic reward fine-tuning process to make a reward pessimistic. Based on PET, we develop a three-step RLHF learning framework as shown in Alg \ref{alg:main}. In this framework, the first step learns a proxy reward model that minimizes the prediction loss. The second step fine-tunes the proxy reward to make it pessimistic. In this setup, the second step is initialized from the proxy reward with a low prediction loss already. This intuitively reduces the complexity of pessimistic reward fine-tuning. The last step optimizes a policy model on the pessimistic reward. In principle, it is unnecessary in the last step to use additional pessimistic learning tricks such as regularization during policy optimization, as the reward model is already pessimistic. The agent can greedily optimize a policy to achieve a high pessimistic reward without worrying about the risk of reward hacking. For example, in the standard KL-PPO algorithm, the optimization goal of the PPO algorithm is the reward of a policy regularized by the KL divergence between the policy and the proxy dataset policy. In our learning framework, the empirical results in Section \ref{sec:exp} show that the PPO algorithm can learn a high-quality policy in reality on the pessimistic reward directly without using any regularization. This empirically verifies that the pessimistic reward fine-tuned by PET is trustworthy.

\begin{algorithm}[tb]
    \caption{Three-Step Pessimistic RLHF Framework}
    \begin{algorithmic}[1]
    \State {\bfseries Input:} Dataset $\mathcal{D}$, Base policy $\pi_0$, Reference policy $\pi_{\text{ref}}$, number of samples $n$ 

    \State {\bfseries Step 1:} Reward modeling $\hat{r}_1 \leftarrow \min_{r\in\mathcal{R}} \mathcal{L}_{\mathcal{D}}(r)$
    
    \State {\bfseries Step 2:} Pessimistic reward fine-tuning $\hat{r}_2 \leftarrow \textbf{PET}(\hat{r}_1)$

    \State {\bfseries Step 3:} Policy optimization $\hat{\pi} \leftarrow \max_{\pi \in \Pi} V^\mu_{\hat{r}_2}(\pi)$
    
    \State {\bfseries Return:} {Policy model $\hat{\pi}$}
    \end{algorithmic}  
    \label{alg:main}
\end{algorithm}

\textbf{Theoretical guarantees:} Here we provide theoretical guarantees on the optimality of the solution to Eq \ref{eq:mini}. The proof technique follows the one in \citet{liu2024provably}. We take a standard definition of the dataset coverage on a policy in Definition \ref{def:2} to characterize how well a dataset covers a policy $\pi$. 

\begin{definition}\label{def:2}
    For a policy $\pi$, given a reference policy $\pi_{\text{ref}}$, a dataset distribution $\mu_{\mathcal{D}}$, and a reward model class $\mathcal{R}$, the coverage coefficient is defined as \cite{zhan2023provable} $\mathcal{C}_{\mu_{\mathcal{D}}}(\mathcal{R},\pi,\pi_{\text{ref}}):=$ $$\max \{0, \sup_{r\in\mathcal{R}}\frac{\mathbb{E}_{x \sim \mu, a_1 \sim \pi(\cdot|x), a_2 \sim \pi_{\text{ref}}(\cdot|x)}[\bigl(r^*(x,a_1)-r^*(x,a_2)\bigr)-\bigl(r(x,a_1)-r(x,a_2)\bigr)]}{\mathbb{E}_{x,a_1,a_2 \sim \mu_{\mathcal{D}}}[|\bigl(r^*(x,a_1)-r^*(x,a_2)\bigr)-\bigl(r(x,a_1)-r(x,a_2)\bigr)|^2]}\},$$
    where $\mu$ is the prompt distribution in $\mu_{\mathcal{D}}$.
\end{definition}

Formally, in Theorem \ref{thm:main}, we show a theoretical guarantee on the performance of the policy in the solution to the minimax problem in Eq \ref{eq:mini}. The proof for Theorem \ref{thm:main} can be found in the appendix.

\begin{theorem}\label{thm:main}
    Consider a bounded reward model class $\mathcal{R}$ such that $\forall r\in\mathcal{R}, r(\cdot,\cdot)\in[-R,R]$ .Assume the true reward is included in the reward model class $r^* \in \mathcal{R}$. Let $\hat{\pi}=\pi_{\text{RS}}(\pi_0,\hat{r},n)$, where $\hat{r}$ is the reward model solution to the minimax problem in Eq \ref{eq:mini}. For any RS policy $\pi \in \Pi^{n,\pi_0}_{\text{RS}}$ covered by the dataset $\mathcal{C}_{\mu}(\pi,\pi_{\text{ref}},\mathcal{R})<+\infty$, by setting $\beta=\frac{\sqrt{N}(1+\exp(R))^2}{2\sqrt{6}\cdot\sqrt{\log(\frac{N_{\epsilon}(\mathcal{R},\|\cdot\|_{\infty})}{\delta})}}$, with probability at least $1-\delta$ the performance gap between $\hat{\pi}$ and $\pi$ is bounded by

    $$V^\mu_{r^*}(\pi)-V^\mu_{r^*}(\hat{\pi}) \leq \frac{(1+\exp(R))^2\cdot(\mathcal{C}_{\mu_{\mathcal{D}}}(\mathcal{R},\pi,\pi_{\text{ref}})^2+1)\cdot \sqrt{6\log(\frac{N_{\epsilon}(\mathcal{R},\|\cdot\|_{\infty})}{\delta})}}{4\sqrt{N}},$$
    where $N_{\epsilon}(\mathcal{R},\|\cdot\|_{\infty})$ is the $\epsilon$-covering number of the reward model class \cite{cheng2022adversarially}.
\end{theorem}

\begin{remark}
    It is standard in the related literature \cite{liu2024provably,zhan2023provable} to assume that the true reward is included in the reward model class. Theorem \ref{thm:main} shows that the performance of $\hat{\pi}$ is comparable to any RS policy $\pi \in \Pi_{\text{RS}}^{\pi_0,n}$ if the policy is covered by the dataset. As long as a high-quality policy is well covered by the dataset, the solution $\hat{\pi}$ will also have high quality, which indicates that reward hacking will not happen. This is an ideal result one can hope for in the offline learning setting \cite{huang2024correcting}.
\end{remark}

\section{Experiments}\label{sec:exp}


\subsection{Experiment Setup}
\textbf{Dataset and model:} Our experiments focus on the standard TL;DR summarization datasets \citep{volske-etal-2017-tl,stiennon2020learning} and use the Pythia-1b model \citep{biderman2023pythiasuiteanalyzinglarge} for all policy and reward models during training. Here, the task is to generate a summary for a given piece of text. We consider the IMDB dataset \citep{maas-etal-2011-learning} as an ablation study. 

\textbf{Implementation:} For the reference policy $\pi_{\text{ref}}$, it is trained by the standard SFT process in \citep{huang2024nimplementationdetailsrlhf}. The base policy $\pi_0$ for the rejection sampling process uses the same SFT model. During inference, the base policy $\pi_0$ uses a higher temperature ($T=0.7$) than the reference policy ($T=0.1$) to increase the randomness in the generation and boost the efficiency of rejection sampling. Due to limitations on our computational resources, we set $n=64$ for rejection sampling during PET, while PET in principle can work with arbitrarily large values of $n$. The proxy reward model is trained by the standard reward modeling process in \citep{huang2024nimplementationdetailsrlhf}. For fair comparison, all RLHF methods share the same reference policy and the proxy reward, if needed. We set the pessimistic coefficient $1/\beta=0.1$ following the setups in \citep{rafailov2023direct}. 

\textbf{Evaluation:} We evaluate the policy models learned by the RLHF algorithms on the test split of the TL;DR SFT dataset, which consists of (query, human summary) pairs. First, we used the policy model to generate a summary for each query in the dataset. Then, we used a judge LLM (Qwen-2.5-32B-Instruct) to evaluate the quality of the generated summaries against the human responses in the dataset. The win rate of the model against the human response is computed by the percentage of instances where the judge LLM preferred the model's summary over the human summary. 

\textbf{Baselines:} We consider multiple state-of-the-art offline RLHF algorithms as baselines, including the traditional PPO-KL \cite{bai2022training}, traditional rejection sampling/best of N sampling \cite{touvron2023llama} (on the proxy reward), DPO \cite{rafailov2023direct}, RPO \citep{liu2024provably}, and $\chi$-PO \citep{huang2024correcting}. Our implementations of DPO and RPO are directly from the source code provided in the original papers. We re-implement the $\chi$-PO algorithm as the source code is unavailable. We use the same hyperparameter setup mentioned in the paper to ensure fair comparison. We try to re-implement another related RLHF baseline, `InferenceTimePessimism' \citep{huang2025best}. The algorithm is based on the rejection sampling process, and the number of samples is set as $n=2^{13}$ in the paper. Due to our computational resource limitation, we set the number of samples as $n=64$. This setup probably underestimates the method's efficiency, and its learned policy has a low win rate. Therefore, we skip this baseline.



\subsection{Pessimistic Reward Finetuning Improves Rejection Sampling}
Here, we evaluate our three-step learning framework with the rejection sampling process for policy optimization. We test the number of samplings to be $n=\{16,32,64,128\}$ during PET and policy optimization, and the results are shown in Table \ref{table:rs}. We observe that the rejection sampling policy based on the pessimistic reward model significantly outperforms the one based on the proxy reward model. The finding empirically validates that the rejection sampling process greatly benefits from the adversarial training process against a reward model.

\begin{table}[!ht]
\centering
\begin{tabular}{@{}l|rrrr@{}}

\hline
 & $n=16$ & $n=32$ & $n=64$ & $n=128$ \\
\hline
\textbf{RS-PET} & $\mathbf{38.2}$ & $\mathbf{36.6}$ & $\mathbf{39.2}$ & $\mathbf{36.8}$\\
RS-Proxy & $28.4$ & $32.0$ & $32.0$ & $34.2$ \\
\hline
\end{tabular}%

\caption{Win rate against human responses for the rejection sampling process on the pessimistic or proxy reward with different numbers $n$ of samplings on the IMDB dataset. `RS-PET' means the rejection sampling process is performed on the pessimistic reward fine-tuned by PET. Similarly, `RS-Proxy' means the reward model is the proxy reward.}
\label{table:rs}
\end{table}

\subsection{KL Regularization Is Unnecessary For Policy Optimization On Pessimistic Reward}
\begin{table}[!ht]
\centering
\begin{tabular}{@{}l|rr|rr@{}}
\hline
 & \multicolumn{2}{|c|}{regularized} & \multicolumn{2}{|c}{\textbf{unregularized}} \\
 & win rate & KL & win rate & KL \\
\hline
\textbf{PPO-Pessimistic} & $36.0$ & $11.6$ & $\mathbf{40.8}$ & $114.0$\\
PPO-Proxy & $\mathbf{40.2}$ & $9.2$ & $7.2$ & $192.6$ \\
\hline
\end{tabular}%
\caption{The policies here are learned by the PPO algorithm on the pessimistic reward or proxy reward, with (`regularized') or without KL (`unregularized') regularization. We show the win rate against human responses of the policies and their KL divergence to the reference policy.}
\label{table:kl}
\end{table}

As introduced earlier, the proxy reward model may overestimate the rewards for out-of-distribution prompt responses, so KL regularization is always considered necessary during policy optimization to prevent reward hacking. In this section, we show that our pessimistic reward model, which is fine-tuned on the proxy reward, is robust against reward hacking such that no regularization is necessary when optimizing a policy on the pessimistic reward.  
Here, we use the PPO algorithm for the policy optimization step in the three-step learning framework. We compare the two cases where KL regularization or no regularization is applied during policy optimization, and we train the algorithms on the proxy reward and the pessimistic reward. Without regularization, the agent would greedily search for the policy with the highest reward. Such an agent is usually considered vulnerable to reward hacking. The results are shown in Table \ref{table:kl}. We observe that for the proxy reward model, the policy learned under the KL regularization performs much better than one with no regularization. The results confirm that the proxy reward model is not pessimistic, and greedily optimizing a policy on proxy reward without KL regularization is vulnerable to reward hacking. In contrast, for the pessimistic reward model, the policy learned under no regularization performs better than the one with regularization. The results intuitively show that the pessimistic fine-tuning successfully makes the reward model pessimistic, and a greedy agent is robust against reward hacking when learning from the pessimistic reward model without using any regularization.

We notice another important finding from Table \ref{table:kl}. Traditionally, the experience is that policies with high KL divergence from the reference policy have a high risk of reward hacking, so they should be excluded during learning through regularization. This agrees with the case of optimizing a policy on the proxy reward with no regularization, as the learned policy has low performance and high KL divergence. However, such an experience does not hold for the case when the reward model is a pessimistic reward. We find that the policy learned on the pessimistic reward fine-tuned by PET has a high performance in reality while having a large KL divergence. This is a sign of over-pessimism when using KL regularization for pessimism, as such good policies are excluded. Our empirical results prove that it is possible to find a high-quality policy on the dataset with a large KL divergence from the reference policy. To include such policies during learning without causing reward hacking, one can perform policy optimization on the pessimistic reward fine-tuned by PET.

\subsection{Comparison between Different Algorithms on Summarization and IMDB Datasets}

Here, we compare our method against the RLHF baselines. We consider using the PPO algorithm for policy optimization with no KL regularization in our three-step learning framework, and we call it `PET-PPO'. Since the RLHF methods we consider generally apply to other NLP tasks in principle, we evaluate them on the IMDB dataset as well to make the evaluation more comprehensive. In this task, the model is prompted with a prefix to a movie review, and it must generate a continuation of that review with positive sentiment. Our data construction and evaluation metric exactly follows the setups in \citet{rafailov2023direct}. To make the task more challenging, the size of our dataset is about $25\%$ of the dataset's size described in \citet{rafailov2023direct}. Training with less data can better highlight the sampling efficiency of an algorithm. The evaluation metric is the likelihood of positive sentiment for the completions generated by the model judged by an LLM. The same LLM is used to generate the preference labels in the dataset, so the likelihood is the ground truth in the task. 

\begin{table}[!ht]
\centering
\begin{tabular}{@{}l|rrrrrrr@{}}
\hline
        &SFT & RS-proxy & KL-PPO & DPO & RPO	& $\chi$PO & \textbf{PET-PPO}\\
        \hline
        Summarization & $25.6$ & $32.0$ & $\mathbf{40.2}$ & $38.8$ & $34.0$ & $39.2$ & $\mathbf{40.8}$ \\ 
        \hline
        IMDB & $89.6$ & $90.7$ & $91.8$ & $95.0$ & $93.5$ & $92.8$ & $\mathbf{98.3}$ \\ 
        \hline
    \end{tabular}
    \caption{Performance of different RLHF methods learning on the summarization and IMDB dataset. The metric on the summarization dataset is the win rate against human responses. On the IMDB dataset, the metric is the likelihood of the responses to have positive sentiment.}
\label{table:win}
\end{table}


In Table \ref{table:win}, we show the standard evaluations on the performance of the policies learned by different algorithms. On the IMDB dataset, our method significantly outperforms the baselines by a clear margin. Note that in the IMDB dataset, a policy model is strictly better than another model if it achieves a higher evaluation score, as the score is given by the ground truth. On the summarization dataset, our method outperforms other methods in general. Here, the win rate is compared against human response, so a model does not necessarily beat another model if it has a higher win rate. Therefore, we directly compare policy models learned by the baselines against our policy model. In Table \ref{table:duel}, we show the win rate of our PPO-PET method against other methods. We observe that, in general, the policy model learned by PPO-PET has a $>50\%$ win rate against other models. The empirical results show that the performance of our PET-PPO method is very competitive compared to current SOTA RLHF methods.

\begin{table}[!ht]
    \centering
    \begin{tabular}{@{}l|rrrrrr@{}}
    \hline
        ~ & SFT & RS-proxy& KL-PPO  & DPO & RPO & $\chi$PO \\ \hline
        \textbf{PET-PPO} & $\mathbf{72.0}$ & $\mathbf{61.6}$ & $\mathbf{57.2}$ & $50.1$ & $\mathbf{58.4}$ & $52.8$\\ \hline

    \end{tabular}
    \caption{Direct comparison between our PET-PPO method against other RLHF baselines on the summarization dataset.}
\label{table:duel}
\end{table}

\section{Related Work}\label{sec:related}
\subsection{RLHF based on explicit reward modeling}
It is typical in RLHF to train an explicit reward model first and then optimize a policy on the learned reward model. A popular choice for policy optimization is the rejection sampling technique \citep{bai2022constitutional,touvron2023llama}. Theoretical understandings of the properties of the rejection sampling process have been developed to explain the efficiency of the method \citep{beirami2024theoretical,huang2025best}. Empirically, different variants of rejection sampling have been proposed to improve the learning efficiency and robustness against reward hacking \citep{huang2025best,jinnai2024regularized,liu2023statistical,xiong2025minimalist,khaki2024rs}. Another popular choice for policy optimization is to use the PPO algorithm \citep{schulman2017proximal}. In this case, it is prevalent in current methods to add a KL regularization in the policy optimization goal to avoid reward hacking \citep{stiennon2020learning,bai2022training,ouyang2022training,christiano2017deep}. To improve the learning efficiency of KL-PPO, methods have been proposed to improve the quality of the reward model \citep{liu2024rrm,sun2025rethinking,shen2024boosting,rame2024warm} or further constrain the agent from uncovered policies \citep{dai2025mitigating}

\subsection{RLHF based on direct preference alignment}
Another typical RLHF technique skips the step of explicit reward modeling through direct preference alignment, also known as direct preference optimization (DPO) \citep{rafailov2023direct}. In DPO, a reward model is implicitly represented by a policy model, so the agent can directly train a policy model to minimize the prediction loss. Numerous direct preference alignment techniques have been proposed to achieve a higher learning efficiency \citep{azar2024general,xiong2023gibbs,tang2024generalized,ji2024towards,liang2024robust,rosset2024direct}. Provably robust algorithms against reward hacking based on DPO have also been developed \citep{fisch2024robust,zhang2024self,xie2024exploratory,liu2024provably,cen2024value, ji2024self}. Note that KL regularization plays a critical role in DPO, as the regularization term makes it possible to represent a reward model by the optimal policy on it under the regularization.

\subsection{RLHF with online preference collection}
This work focuses on RLHF in the offline setting, where the learning agent only has access to an offline preference dataset. Another important setting of RLHF is the online setting, where an agent can collect preference data online. The online setting can break the limit of the offline setting in that the agent can control the data distribution to cover high-quality policies actively. The online setting is also more expensive than the offline setting, as it needs to collect customized data during training. Theoretical studies have developed efficient online exploration algorithms to improve the quality of online data collection \cite{qi2025sample,wu2023making,zhao2025logarithmic}. Practical methods and learning frameworks have also been developed and have achieved promising empirical results \cite{dong2024rlhf,xiong2023iterative,bai2022training,touvron2023llama}. 
\section{Conclusions and Limitations}
This work develops PET, a reward fine-tuning method. A pessimistic reward model can be learned by fine-tuning a proxy reward with PET. When optimizing a policy on a pessimistic reward, a greedy agent can learn a high-performing policy with no regularization. We develop an RLHF method called PET-PPO that uses the PPO algorithm for policy optimization on a pessimistic reward fine-tuned by PET. Our empirical results show that PET-PPO achieves comparable or stronger performance on the IMDB and the TL;DR summarization dataset than multiple current SOTA RLHF baselines. The scope of this work is limited to RLHF in the offline setting. Due to limited computational resources, the evaluation is performed on a subset of RLHF tasks based on LLMs with relatively small size.  

\bibliography{main.bib}
\bibliographystyle{plainnat}

\newpage

\appendix 
\section{Detailed Theoretical Analysis}
\subsection{Theoretical understanding on solving the minimax problem in Eq \ref{eq:mini} with PET}

Here, we explain in detail how the PET algorithm in Alg \ref{alg:advbon} solves the theoretical problem in Eq. \ref{eq:mini}. Recall that Eq. \ref{eq:mini} is a minimax problem based on rejection sampling as follows:

\begin{equation*}
    \min_{r \in \mathcal{R}}\max_{\pi \in \Pi^{n,\pi_0}_{\text{RS}}} \bigl(V^\mu_r(\pi)-V^\mu_r(\pi_{\text{ref}})\bigr)+\beta \cdot \mathcal{L}_{\mathcal{D}}(r).
\end{equation*}

By Proposition \ref{pro:rs}, we have $\pi_{\text{RS}}(\pi_0,n,r^t)\in\arg\max_{\pi\in\Pi^{n,\pi_0}_{\text{RS}}}V^\mu_r(\pi).$ Therefore, solving the minimax problem in Eq \ref{eq:mini} is equivalent to solving the minimization problem as follows:

\begin{equation*}
    \min_{r \in \mathcal{R}}\bigl(V^\mu_r(\pi_{\text{RS}}(\pi_0,n,r))-V^\mu_r(\pi_{\text{ref}})\bigr)+\beta \cdot \mathcal{L}_{\mathcal{D}}(r).
\end{equation*}

Denote $$f(r,\pi):=\bigl(V^\mu_r(\pi)-V^\mu_r(\pi_{\text{ref}})\bigr)+\beta \cdot \mathcal{L}_{\mathcal{D}}(r)$$ and $$h(r):=\bigl(V^\mu_r(\pi_{\text{RS}}(\pi_0,n,r^t))-V^\mu_r(\pi_{\text{ref}})\bigr)+\beta \cdot \mathcal{L}_{\mathcal{D}}(r).$$


Then, our goal is to solve the problem $\min_{r\in\mathcal{R}}h(r)$. For any reward model $r_0 \in \mathcal{R}$, we have 

\begin{equation*}
    \begin{aligned}
        \nabla h(r)|_{r=r_0} &= \nabla_r f(r,\pi)|_{r=r_0,\pi=\pi_{\text{RS}}(\pi_0,n,r_0)}+ \nabla_{\pi} f(r,\pi)|_{r=r_0,\pi=\pi_{\text{RS}}(\pi_0,n,r_0)}\cdot\nabla_r \pi_{\text{RS}}(\pi_0,n,r_0)|_{r=r_0}\\
        &= \nabla_r f(r,\pi)|_{r=r_0,\pi=\pi_{\text{RS}}(\pi_0,n,r_0)}
    \end{aligned}
\end{equation*}

The second equality holds because $\pi=\pi_{\text{RS}}(\pi_0,n,r)$ is the optimizer for $\max_{\pi\in\Pi^{n,\pi_0}_{\text{RS}}}f(r,\pi),$ so we have $\nabla_{\pi} f(r,\pi)|_{r=r_0,\pi=\pi_{\text{RS}}(\pi_0,n,r_0)}\equiv 0,\forall r_0\in\mathcal{R}$. This is an important result as it shows that to compute $\nabla h(r)|_{r=r_0}$, we don't need to compute $\nabla_r \pi_{\text{RS}}(\pi_0,n,r)|_{r=r_0}.$ The latter represents how the rejection sampling would change if the reward function changes, which is hard to approximate in practice. 

A standard approach to find the minimizer of $\min_{r}h(r)$ is to perform stochastic gradient descent on $h(r).$ By the previous results, we can implement this process by approximating the value of $\nabla h(r)|_{r=r_0}=\nabla_r f(r,\pi)|_{\pi=\pi_{\text{RS}}(\pi_0,n,r^t))}.$ 

Next, we show that PET in Alg \ref{alg:advbon} is essentially the process of stochastic gradient descent to solve the problem $\min_{r\in\mathcal{R}} h(r)$. Recall $V^r_\mu(\pi)=\mathbb{E}_{x\sim \mu,a\sim \pi(\cdot|x)}[r(x, a)].$ At line $5$, the prompts $x_i$ and the data minibatch $\mathcal{D}_t$ are sampled from the dataset $\mathcal{D}$. At line $6$, the responses $a_i,a_{\text{ref},i}$ are stochastically sampled from the current policy $\pi^t$ and the reference policy $\pi_{\text{ref}}.$ Therefore, the expectation of the average loss $l^t/M$ equals to the value of $h(r^t)$:

$$\mathbb{E}[\sum_{i\in[M]} [r^t(x_i,\pi^t(x^i,a_i))-r^t(x_i,\pi^t(x^i,a_{\text{ref},i}))+\beta \cdot \mathcal{L}_{\mathcal{D}_t}(r^t)]]/M=\bigl(V^\mu_r(\pi^t)-V^\mu_r(\pi_{\text{ref}})\bigr)+\beta \cdot \mathcal{L}_{\mathcal{D}}(r)$$ 

At line $8$, the algorithm computes the gradient of $l^t$, so Alg \ref{alg:advbon} is essentially a standard process of stochastic gradient descent to solve the minimization problem $\min_{r\in\mathcal{R}}h(r).$ Therefore, PET is a convenient implementation to solve the minimax problem based on rejection sampling in Eq. \ref{eq:mini}.

\begin{remark}
    Here we argue that PET approximates the most pessimistic low-prediction-loss reward model against the rejection sampling process. Denote $\hat{r}$ the solution to the minimization problem $\hat{r}\in\min_{r\in\mathcal{R}}\bigl(V^\mu_r(\pi)-V^\mu_r(\pi_{\text{ref}})\bigr)+\beta \cdot \mathcal{L}_{\mathcal{D}}(r).$ For any reward model $r \in \mathcal{R}$ that has the same or lower prediction loss on the dataset $\mathcal{L}_{\mathcal{D}}(r) \leq \mathcal{L}_{\mathcal{D}}(\hat{r})$, we have $\bigl(V^\mu_{\hat{r}}(\pi)-V^\mu_{\hat{r}}(\pi_{\text{ref}})\bigr)\leq \bigl(V^\mu_{r}(\pi)-V^\mu_{r}(\pi_{\text{ref}})\bigr).$ This is because
    \begin{equation*}
        \begin{aligned}
            &\bigl(V^\mu_{\hat{r}}(\pi)-V^\mu_{\hat{r}}(\pi_{\text{ref}})\bigr)+\beta \cdot \mathcal{L}_{\mathcal{D}}(r) \leq \bigl(V^\mu_r(\pi)-V^\mu_r(\pi_{\text{ref}})\bigr)+\beta \cdot \mathcal{L}_{\mathcal{D}}(r)\\
            \Rightarrow&\bigl(V^\mu_{\hat{r}}(\pi)-V^\mu_{\hat{r}}(\pi_{\text{ref}})\bigr) - \bigl(V^\mu_r(\pi)-V^\mu_r(\pi_{\text{ref}})\bigr) \leq \beta \cdot (\mathcal{L}_{\mathcal{D}}(r)-\mathcal{L}_{\mathcal{D}}(\hat{r}))\\
            \Rightarrow&\bigl(V^\mu_{\hat{r}}(\pi)-V^\mu_{\hat{r}}(\pi_{\text{ref}})\bigr) - \bigl(V^\mu_r(\pi)-V^\mu_r(\pi_{\text{ref}})\bigr) \leq 0\\
            \Rightarrow&\bigl(V^\mu_{\hat{r}}(\pi)-V^\mu_{\hat{r}}(\pi_{\text{ref}})\bigr) \leq \bigl(V^\mu_r(\pi)-V^\mu_r(\pi_{\text{ref}})\bigr)
        \end{aligned}
    \end{equation*}
    In practice, we use the PET algorithm to approximate the solution $\pi_0$ and find that the prediction loss of the learned reward model is as low as that of the proxy reward, which is specially trained to minimize the prediction loss. This implies that the reward model $\hat{r}$ is also the most pessimistic reward model that gives minimal relative score to the rejection sampling process among the reward models with low values of prediction loss. This observation also supports our argument that the reward model learned from PET is pessimistic.
\end{remark}

\subsection{Proof for Proposition \ref{pro:rs}}

Proposition \ref{pro:rs} describes an intuitive property of rejection sampling. It is equivalent to the statement that when optimizing a policy on a reward model $r_1$ through the rejection sampling process, the RS process that sets its reward model as $r=r_1$ achieves the highest reward on $r_1$. For any rejection sampling process with the same base policy and number of samples, the distribution of the sampled responses during the process is always the same. Therefore, for any outcome of sampled responses, the RS process that sets $r=r_1$ can always output the response with the highest reward on $r_1$. Formally, the proof for Proposition \ref{pro:rs} is as follows.

\begin{proof}
    Given a base policy $\pi_0$, a positive integer $n$, and a prompt $x$, consider a stochastic process where $n$ responses are i.i.d drawn from the policy at the prompt $a_i \overset{i.i.d}{\sim} \pi_0(\cdot|x),i\in[n].$ Let $E(x)$ be the space of all possible outcomes. For any outcome $e \in E(x)$, let $\{a_1,\ldots,a_n\}$ be the sampled responses. For any two different reward models $r_1\neq r_2$, denote $i^*_1 \in \arg\max_{i\in[n]}r_1(x,a_i)$ and $i^*_2 \in \arg\max_{i\in[n]}r_2(x,a_i)$ as the optimal index on the two rewards. Denote $v_1(e,r):=r(x,a_{i^*_1})$ and $v_2(e,r):=r(x,a_{i^*_2})$ where $r$ is any reward model, then by definition, we have $v_1(e,r_1) \geq v_2(e,r_1).$ Consider two rejection sampling policies $\pi_1=\pi_{\text{RS}}(\pi_0,r_1,n), \pi_2=\pi_{\text{RS}}(\pi_0,r_2,n)$ defined on the base policy $\pi_0$, sampling number $n$, and the reward models $r_1,r_2$. Their performance on the reward model $r_1$ for any prompt distribution $\mu$ satisfies $V^\mu_{r_1}(\pi_1)=\mathbb{E}_{x\sim \mu, e\sim E(x)}[v_1(e,r_1)]$, $V^\mu_{r_1}(\pi_2)=\mathbb{E}_{x\sim \mu, e\sim E(x)}[v_2(e,r_1)]$. By the previous result $v_1(e,r_1) \geq v_2(e,r_1),$ we have $V^\mu_{r_1}(\pi_1) \geq V^\mu_{r_1}(\pi_2),$ which conclude the proof.
    
\end{proof}

\subsection{Proof for Theorem \ref{thm:main}}
\begin{proof}
Our proof generally follows the proof technique in \citet{liu2024provably}. Let $\hat{r}$ be the reward solution to $\hat{r} \in \arg\min_{r\in\mathcal{R}}$
\begin{equation*}
    \begin{aligned}
        &V^\mu_{r^*}(\pi)-V^\mu_{r^*}(\hat{\pi}) \\
        &= \bigl(V^\mu_{r^*}(\pi) - V^\mu_{\hat{r}}(\pi)\bigr)+\bigl(V^\mu_{\hat{r}}(\pi) - V^\mu_{\hat{r}}(\hat{\pi})\bigr)+\bigl(V^\mu_{\hat{r}}(\hat{\pi}) - V^\mu_{r^*}(\hat{\pi})\bigr) \\
        &\leq \bigl(V^\mu_{r^*}(\pi) - V^\mu_{\hat{r}}(\pi)\bigr) + \bigl(V^\mu_{\hat{r}}(\hat{\pi}) - V^\mu_{r^*}(\hat{\pi})\bigr) \\
        &= \bigl(V^\mu_{r^*}(\pi) - V^\mu_{\hat{r}}(\pi)\bigr) + \bigl(V^\mu_{\hat{r}}(\hat{\pi}) -V^\mu_{\hat{r}}(\pi_{\text{ref}}) + \beta \cdot \mathcal{L}_{\mathcal{D}}(\hat{r})  - (V^\mu_{r^*}(\hat{\pi}) - V^\mu_{\hat{r}}(\pi_{\text{ref}}) + \beta \cdot \mathcal{L}_{\mathcal{D}}(r^*))\bigr) +\\
        & \beta \cdot (\mathcal{L}_{\mathcal{D}}(r^*)-\mathcal{L}_{\mathcal{D}}(\hat{r}))\\
        &\leq \bigl(V^\mu_{r^*}(\pi) - V^\mu_{\hat{r}}(\pi)\bigr) + \beta \cdot (\mathcal{L}_{\mathcal{D}}(r^*)-\mathcal{L}_{\mathcal{D}}(\hat{r})) \\
    \end{aligned}
\end{equation*}

The first equality utilizes the optimality of the policy solution $\hat{\pi} \in \arg\max_{\pi \in \Pi_{\text{RS}}} V^\mu_{\hat{r}}(\pi)$. The second equality utilizes the optimality of the reward solution $\hat{r} \in \arg\min_{r\in\mathcal{R}} \bigl(V^\mu_{\hat{r}}(\pi)-V^\mu_{\hat{r}}(\pi_{\text{ref}})+\beta\cdot\mathcal{L}_{\mathcal{D}}(r)\bigr)$. Intuitively, the formulation in the last line is bounded for any reward function $\hat{r}$. If the reward function is close to the true reward $r^*$, then both differences $V^\mu_{r^*}(\pi) - V^\mu_{\hat{r}}(\pi)$ and $\mathcal{L}_{\mathcal{D}}(r^*)-\mathcal{L}_{\mathcal{D}}(\hat{r})$ should be small. If $\hat{r}$ is very different from $r^*$, then its prediction loss on the dataset $\mathcal{L}_{\mathcal{D}}(\hat{r})$ would also be high, so that the performance gap can still be bounded. This indicates the importance of adding the prediction loss as a constraint on the reward model in Eq. \ref{eq:mini}. Formally, based on the results of Theorem 5.3 in \citet{liu2024provably}, with probability at least $1-\delta$, the last term can be bound by 

$$V^\mu_{r^*}(\pi)-V^\mu_{r^*}(\hat{\pi})\leq \frac{\mathcal{C}_{\mu_{\mathcal{D}}}(\mathcal{R},\pi,\pi_{\text{ref}})^2}{8 \kappa^2 \cdot \beta}+\frac{3\beta}{N}\log(\frac{N_{\epsilon}(\mathcal{R},\|\cdot\|_{\infty})}{\delta}).$$

Here, $\kappa=\frac{1}{(1+\exp(R))^2}$ is a constant, and $N_{\epsilon}(\mathcal{R},\|\cdot\|_{\infty})$ is the $\epsilon$-covering number for the reward model class \cite{cheng2022adversarially}. Setting $$\beta =\frac{\sqrt{N}}{2\kappa \cdot\sqrt{6\log(\frac{N_{\epsilon}(\mathcal{R},\|\cdot\|_{\infty})}{\delta})}}$$ concludes the proof. Note that the bound here is tighter than the bound on \citet{liu2024provably}, because there is no KL regularization in our formulation at Eq. \ref{eq:mini}. Adding a KL regularization multiplied by any positive coefficient to Eq. \ref{eq:mini} will increase the bound on the performance gap analysis, which is undesired for sampling efficiency.
\end{proof}

\section{Policy Optimization Methods}
\textbf{KL regularized proximal policy optimization (KL-PPO):}
For a prompt distribution $\mu$, the `KL divergence' between two policies can be defined as $\text{KL}_{\mu}(\pi_1,\pi_2):=\mathbb{E}_{x\sim\mu}[\text{KL}(\pi_1(\cdot|x)\|\pi_2(\cdot|x))]$, where $\text{KL}(\pi_1(\cdot|x)\|\pi_2(\cdot|x)):=\sum_{a\in\mathcal{A}}\pi_1(a|x)\cdot \log \frac{ \pi_1(a|x)}{\pi_2(a|x)}$ is the standard KL divergence between two distributions. Consider a proxy policy $\pi_{\text{ref}}$ of the dataset. The proxy policy is trained to generate a response distribution that is similar to the dataset response distribution. This is often achieved by a standard supervised fine-tuning process on the prompts and responses from the dataset $\mathcal{D}$ \citep{touvron2023llama}. The KL divergence between a policy $\pi$ and $\pi_{\text{ref}}$ intuitively indicates how well it is covered by the dataset. Then, the learning goal of the agent becomes $$\hat{\pi}\leftarrow \arg\max_{\pi \in \Pi} V^{\mu}_{\hat{r}}(\pi)+\eta \cdot \text{KL}_\mu(\pi,\pi_{\text{ref}}),$$ where $\Pi$ is a model family, $\eta>0$ is the weight of the KL regularization in the optimization target. The most popular way to solve this optimization problem is by using the PPO algorithm \citep{ouyang2022training}. Combining the reward modeling step and the policy optimization process gives the popular RLHF algorithm `PPO-KL' as shown in Alg \ref{alg:ppo}

\textbf{Rejection sampling process (RS):} Rejection sampling, also known as best-of-N sampling, is an inference-time policy optimization method. Given a base policy model $\pi_0$, a reward model $\hat{r}$, and a positive integer $n$, the process of rejection sampling is defined in Alg \ref{alg:bon}. In practice, the reward model $\hat{r}$ is trained by minimizing the prediction loss, and the base policy $\pi_0$ is usually set as a proxy policy for the dataset. The rejection sampling process is effectively a policy as it takes a prompt as input and stochastically outputs a response. An RS policy can be directly implemented on a policy model and a reward model without any training. It also achieves a higher reward on the reward model $\hat{r}$ compared to the base policy $\pi_0$. The pessimism in rejection sampling implicitly relies on the notion of KL regularization. When the value of $n$ is small, the RS policies will have a limited KL regularization compared to the base policy $\pi_0$, which intuitively reduces the risk of reward hacking \cite{beirami2024theoretical}. One can achieve a higher performance on the reward model with rejection sampling by increasing the value of $n$, but in this case, the risk of reward hacking also increases \cite{huang2025best}. In practice, reward hacking has been observed in RS with a relatively high value of $n$ \cite{touvron2023llama,gao2023scaling}. Therefore, it is important to develop principled methods to free RS policies from reward hacking. 

\begin{algorithm}[!ht]
\caption{KL-PPO}
\begin{algorithmic}[1]
    \State {\bfseries Input:} {Reference policy $\pi_{\text{ref}}$, Dataset $\mathcal{D}$, KL weight $\eta$}
    \State {\bfseries Step 1:} Reward modeling $\hat{r} \leftarrow \min_{r\in\mathcal{R}} \mathcal{L}_{\mathcal{D}}(r)$
    
    \State {\bfseries Step 2:} Policy optimization with PPO $\hat{\pi} \leftarrow \max_{\pi \in \Pi} V^\mu_{\hat{r}}(\pi)-\eta\cdot \text{KL}_{\mu}(\pi,\pi_{\text{ref}})$
    
    \State {\bfseries Return:} {Policy model $\hat{\pi}$}

\end{algorithmic}   
\label{alg:ppo}
\end{algorithm}

\begin{algorithm}[!ht]
\caption{Rejection sampling process}
\begin{algorithmic}[1]
    \State {\bfseries Input:} {Base policy $\pi_0$, reward model $\hat{r}$, number of samples $n$, prompt $x$}

    \State Draw responses $a_i \overset{i.i.d.}{\sim} \pi_0(\cdot|x),\forall i\in [n]$
    
    \State Generate rewards $r_i = \hat{r}(x,a_i),\forall i \in [n]$

    \State {\bfseries Return:} {Response $a_{i^*}$, where $i^* \in \arg\max_{i\in[n]} r_i$}
\end{algorithmic}   
\label{alg:bon}
\end{algorithm}

\section{Additional Experiment Details}
\textbf{Hyper-parameter setups:}
\begin{table}[H]
    \centering
    \begin{tabular}{c |cc}
    \toprule 
        Configuration & PET & PPO \\
        \midrule
        learning rate & $3e-8$ & $3e-6$ \\
        learning scheduler type& cosine & cosine \\
        batch size &$128$ &$128$\\
        gradient accumulation steps & 16 & 16\\
        training epoch & $1$ & $1$
        \\
        pessimistic coefficient $\beta$ & $10$ & / \\
        KL regularization weight $\eta$ & / & $0.05$ \\
        optimizer & adamw torch & adamw torch \\
        precision & bfloat16 & bfloat16
        \\
         \bottomrule
    \end{tabular}
    \caption{Training configurations for PET and PPO.} 
    \label{tab:hype}
\end{table}
We list detailed training configurations for PET and PPO in Table \ref{tab:hype}. For standard SFT model and proxy reward model training, we follow the same setup as in \citep{huang2024nimplementationdetailsrlhf}. For all baseline methods, we follow the original authors' implementations or descriptions. Specifically, for RPO, we remove the default chat template provided by its source code, as it is unsuitable for our summarization and IMDB tasks. This is confirmed by our observation that including the template degrades RPO's performance on our tasks, so the template is not included in our RPO training.

\textbf{Statistical significance:}
To verify that the methods we test with have stable output, we test our method PET-PPO and other baselines, including KL-PPO, DPO, RPO, and $\chi$PO, on both TL;DR summarization and IMDB datasets with a different random seed. In Table. \ref{apptable:win} and Table. \ref{apptable:duel}, we show the empirical mean and confidence intervals of the evaluation results on the RLHF methods from the two random seeds.

\begin{table}[!ht]
\centering
\begin{tabular}{@{}l|rrrrr@{}}
\hline
       & KL-PPO & DPO & RPO & $\chi$PO & \textbf{PET-PPO}\\
        \hline
        Summarization & $40.5\pm0.3$ & $37.5\pm1.3$ & $33.6\pm0.4$ & $38.2\pm1.0$ & $\mathbf{41.4\pm0.6}$ \\ 
        \hline
        IMDB & $92.0\pm0.2$ & $\mathbf{95.0\pm0.0}$ & $93.1\pm0.4$ & $93.0\pm0.2$ & $\mathbf{96.1\pm2.2}$ \\ 
        \hline
    \end{tabular}
    \caption{Empirical mean and standard errors of the evaluation results for the RLHF methods on the summarization and IMDB dataset}
\label{apptable:win}
\end{table}

\begin{table}[!ht]
    \centering
    \begin{tabular}{@{}l|rrrr@{}}
    \hline
        ~ & KL-PPO  & DPO & RPO & $\chi$PO \\ \hline
        \textbf{PET-PPO} & $\mathbf{57.2\pm0.0}$ & $51.2\pm1.2$ & $\mathbf{62.0\pm3.6}$ & $\mathbf{55.0\pm2.2}$ \\ \hline

    \end{tabular}
    \caption{Empirical mean and standard errors for the win rate of the PET-PPO method against other RLHF baselines on the summarization dataset.}
\label{apptable:duel}
\end{table}

\textbf{Computational resource:} We use 2 NVIDIA A100-PCIE GPUs, each with 40GB VRAM and 2 48-core230 Intel Xeon Silver 4214R CPU.



\end{document}